\documentclass{ecai}
\usepackage{times}
\usepackage{graphicx}
\usepackage{latexsym}
\usepackage{caption}

\usepackage{amsmath,amssymb,amsfonts}

\begin{document}

\title{DRNet: Dissect and Reconstruct the Convolutional Neural Network via Interpretable Manners}

\author{Xiaolong Hu\institute{Institute of Computing Technology, Chinese Academy of Sciences, China, email: \{huxiaolong18g, anzhulin, yangchuanguang, zhuhui, xukaiqiang, xyj\}@ict.ac.cn} \textsuperscript{\space \rm,}\institute{University of Chinese Academy of Sciences, China\newline*Corresponding author} \space \and Zhulin An\textsuperscript{\rm 1,*} \and Chuanguang Yang\textsuperscript{\rm 1,2} \and Hui Zhu\textsuperscript{\rm 1,2} \and Kaiqiang Xu\textsuperscript{\rm 1,2} \\ \and Yongjun Xu\textsuperscript{\rm 1}}

\maketitle
\bibliographystyle{ecai}

\begin{abstract}

Convolutional neural networks (ConvNets) are widely used in real life. People usually use ConvNets which pre-trained on a fixed number of classes. However, for different application scenarios, we usually do not need all of the classes, which means ConvNets are redundant when dealing with these tasks. This paper focuses on the redundancy of ConvNet channels. We proposed a novel idea: using an interpretable manner to find the most important channels for every single class (dissect), and dynamically run channels according to classes in need (reconstruct). For VGG16 pre-trained on CIFAR-10, we only run 11\% parameters for two-classes sub-tasks on average with negligible accuracy loss. For VGG16 pre-trained on ImageNet, our method averagely gains 14.29\% accuracy promotion for two-classes sub-tasks. In addition, analysis show that our method captures some semantic meanings of channels, and uses the context information more targeted for sub-tasks of ConvNets.

\end{abstract}

\section{Introduction}

Convolutional neural networks (ConvNets) have been broadly used on various visual tasks due to their superior performance\cite{vgg}\cite{resnet}\cite{densenet}. Most of ConvNets are usually pre-trained on a fixed number of classes. However, people usually do not need ConvNet's discrimination ability over all of these classes. This phenomenon tells us that there are resources waste during the inference stage of ConvNets. This kind of waste is tolerable in cloud side. But when it comes to edge devices, even a small improvement in resources saving can help a lot to improve user experiences. Some works have been done to prune neural networks into smaller ones\cite{slimming}\cite{pruning1}\cite{pruning2}. In addition, there are also many light weight network structures proposed to adapt ConvNets to computational limited mobile devices\cite{mobilenet}\cite{mobilenetv2}\cite{shufflenet}\cite{hcg}. However, people still use full networks for inference. To the best of our knowledge, nobody has ever made use of the class set information of a sub-task. A sub-task means people only need to classify a subset of classes. i.e. one sub-task is to classify cats and dogs, and the other sub-task is to classify apples and watermelons. If we have a ConvNet which is pre-trained on ImageNet, we must run the whole ConvNet for each sub-task. Actually, this is unnecessary, because we already know there are only cats and dogs in the first sub-task, apples and watermelons in the second sub-task.

Our work focuses on a basic problem, i.e. can we run only part of a ConvNet channels for a sub-task? To achieve this goal, we need a method to find the important channels for every sub-tasks. However, there are 1024 ($2^{10}$) kinds of class combination for a ConvNet pre-trained on a dataset with 10 classes. For 1000 classes dataset, the combination number is $1.07 \times 10^{31}$ ($2^{1000}$). It will be a huge burden if we try to find important channels for all combinations one by one. To solve this problem, we propose a two-stage method. Firstly, we use an interpretable manner to find important channels for every single class. This process actually extracts some semantic information of each channel. Secondly, we design two strategies to combine the important channels for any sub-task. Then we run sub-tasks only on important channels. Using this method, inference devices only need to store one full ConvNet and the channel importance information for every class. When new sub-tasks come, these inference devices can quickly decide which channels to run, rather than run the whole ConvNet for every sub-task.

\begin{figure}[t]
\centerline{\includegraphics[width=3.0in]{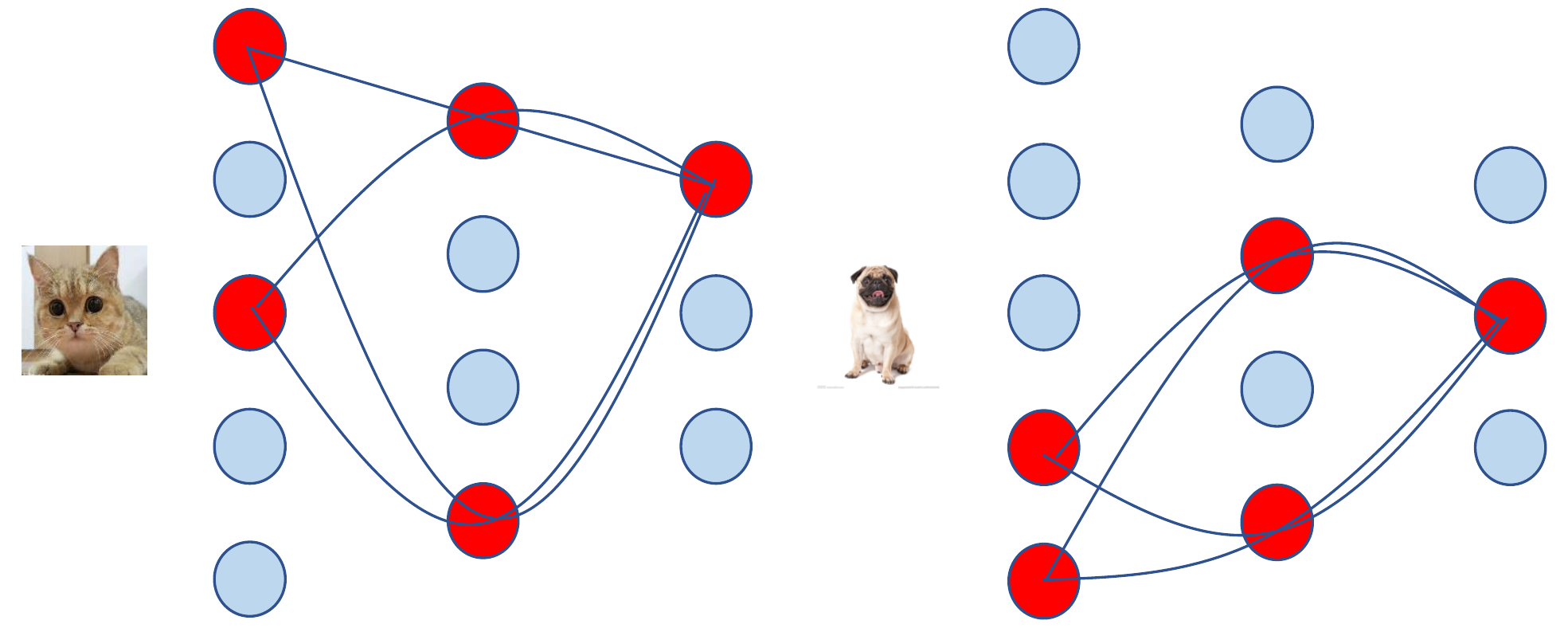}}
\caption{Important channels for cats and dogs. Traditional ConvNets are pre-trained on a fixed number of classes. However, a class do not need all channels to capture its features. This figure shows that different classes have different important channels. Circles in red represent the important channels.} \label{procstructfig}
\end{figure}

Our method is inspired by an interpretable method of ConvNet\cite{cdrp}. As shown in Figure 1, the original ConvNet has several channels, but not every channel is useful for the discrimination of every class. We need to find the important channels for every class and combine them to get the important channels for sub-tasks. Note that we do not mean to propose a pruning method to make ConvNets slimmer, but to run different channels for different sub-tasks. Experiments show that our method not only can be used on large-scale ConvNets such as VGG and ResNet, but also can be used on some light weight structures such as MobileNetV2. In addition, there is nearly no additional computation burden during the inference period.

\subsection{Method}

Our method has two stages: dissection and reconstruction. Suppose that we have a pre-trained ConvNet, at the dissection stage, we extract important channels for every class. At the reconstruction stage, we deal with a certain sub-task. For any sub-task, we combine the important channels of its classes, and dynamically run different channels for different sub-tasks. As the reconstruction process is very efficient, we can easily switch from one sub-task to the others.

We will elaborate our method in Section 3 in detail. 

\subsection{Contribution}
In this paper, our work mainly contributes in the following aspects:

\begin{itemize}
	\item For the dissection stage, we propose to use an interpretable method to extract important channels for every class. For the reconstruction stage, we propose two efficient methods to combine the important channels of several classes.
	\item We dynamically switch the running channels of ConvNets to any sub-task. For VGG16 pre-trained on CIFAR-10, we only run 11\% parameters for two-classes sub-tasks on average with negligible accuracy loss. For VGG16 pre-trained on ImageNet, our method averagely gains 14.29\% accuracy promotion for two-classes sub-tasks.
	\item We do experiments to explain why our methods work. The results show that our method captures some semantic meanings of channels and makes better use of context features.
\end{itemize}

\section{Related Work}

\subsection{Dynamic Inference}
There exists some dynamic inference networks. Wang et al. proposed to skip some layers during the inference process of a ConvNet\cite{skipnet}. They used a self-adaptive method to skip some of the middle layers to save computation resources when making predictions. Surat et al. proposed a method to exit early during inference time\cite{branchynet}. They found that they only need to use the feature maps of front layers to make prediction for most of the test samples. So they designed an algorithm to exit from the front layers if the confidence coefficient of predictions made by the front layers was high enough. Yu et al. used a novel training method to make ConvNet slimmable during inference time\cite{slimmable}\cite{uni-slimmable}.

However, these works are still designed for full-task of ConvNets. Our motivation is to dynamically run part of a ConvNet for any sub-task, which is very different from the previous dynamic inference networks.

\subsection{Network Pruning}
Network pruning technologies have become very mature and have been used in many areas. Un-structured pruning means to dissect convolution kernels into neurals and perform pruning on neural level. Han et al. proposed to prune the unimportant connections with small weights in pre-trained ConvNet\cite{pruning1}. Yang et al. used evolution methods to perform pruning\cite{ga_pruning}. Structured pruning suggests to prune the network on channel level\cite{slimming}\cite{pruning3}\cite{pruning4}. The most typical work was done by Liu et al. They added channel scaling factors after each layer and trained the network with the scaling factors, then removed the channels with low factor values\cite{slimming}.

Network pruning skills can make ConvNet slimmer and shallower. But after they have done the pruning, they still use the whole network for inference. Different from the previous work, our method allows people to dynamically run part of a ConvNet during inference process.

\subsection{Interpretable Manners of ConvNets}
The interpretability of ConvNets has drawn much attention in recent years. Many researchers use visualization methods to interpret how ConvNets work\cite{visualizing1}\cite{visualizing2}\cite{visualizing3}. Zhou et al. addressed the balance between interpretability and performance and proposed a method to measure the interpretability quantitatively\cite{dissection}. Guillaume et al. suggested to use a linear classifier probe to measure the discrimination ability of every layer in ConvNets\cite{linear_probes}.

Zhang et al. designed an interpretable ConvNet and then used an decision tree to interpret the semantic meanings of channels\cite{interpretableCNN}\cite{decision_tree}. Squeeze-and-Excitation networks added a bottleneck structure after every layer, and automatically gave weights to every channel\cite{SE}. Li et al. proposed to use the SE block to measure the importance of channels\cite{decoupling}. Wang and Su proposed the CDRP method to interpret ConvNet\cite{cdrp}. Qiu et al. proposed to dissect channels into neural, and extracted neural path according to their activation values to do adversarial sample detection\cite{path_extraction}.

\section{Method}

As the previous works can not satisfy our needs, we propose a new method named DRNet. In general, our method consists of two stages: dissection and reconstruction. We will elaborate these two stages in Section 3.1 and 3.2.

\subsection{Dissection Stage}

In this section, we present a method to extract the important channels for each class, and this process is called dissection. For every class, we use a float value (in$[0, 1]$) to represents the importance of a single channel. As shown in Figure 2, the vector of importance values is called Channel Importance Vector (CIV). The method is inspired by Critical Data Routine Path (CDRP) proposed in \cite{cdrp}. We will briefly introduce the CDRP method, and explain how we use it to generate CIVs.

In general, the first step of CDRP method is to normally train a traditional ConvNet, and add a control gate after the ReLU operation of each channel in all convolution layers. For each channel, the control gate is a float variable, and it is straightly multiplied to the output feature map. After that, we need to freeze all other parameters and train the control gates only. For each image, we train the network to get a vector of control gates, which is called CDRP. That is to say, every single image has its own CDRP vector. At the last step, we use each value in CDRP to represent the importance of each channel.

In detail, suppose that we have a pre-trained ConvNet $f_\theta$. Each convolutional layer of ConvNet consists of many channels. We put a control gate after the ReLU activation of each channel in all layers.

During the forward propagation, every element of the output feature map simply multiply the control gate. We use $\lambda$ to represent control gates parameters. Every time, we input one image $x$ to the ConvNet.

In order to perform backward propagation, we define the loss function, which is inspired by knowledge distillation\cite{KD}. It can be written as the following form:

\begin{equation}
\mathop{min}\limits_{\lambda}(L(f_\theta(x), f_\theta(x;\lambda)) + \gamma \left \| \lambda \right \|_1)
\label{thesystem}
\end{equation}

$f_\theta(x)$ represents the original network's output and  $f_\theta (x;\lambda)$ represents the output of network with $\lambda$. The first term $L(f_\theta(x), f_\theta(x;\lambda))$ represents the KL loss between the original ConvNet's output and the output after adding the control gates. This loss is also called soft-target loss indicating that the network with control gates tries to output exactly the same as the original network.

The second term $\gamma \left \| \lambda \right \|_1$ is the $L_1$ loss of $\lambda$, which makes $\lambda$ a sparse vector. $\gamma$ is the weight of the $L_1$ loss, and we use it to control the trade-off between two terms.

During the backward propagation, we freeze all parameters in the network except $\lambda$. Formula 2 shows how to calculate the gradient of $\lambda$.

\begin{equation}
\frac{\partial Loss}{\lambda} = \frac{\partial L}{\partial \lambda} + \gamma * sign(\lambda)
\end{equation}

The gradient of $\lambda$ consists of two terms. The first term $\frac{\partial L}{\partial \lambda}$ comes from the soft-target loss. The second term $\gamma * sign(\lambda)$ is the $L_1$ loss.

By using this method, we get a CDRP vector $\lambda_i$ for each image $i$. Then we use the average value of CDRPs in one class to generate the CIV of that class. We use $\lambda^c$ to represent the CIV of class $c$. Figure 2 shows an example of CIV.

\begin{figure}[t]
\centerline{\includegraphics[width=3.0in]{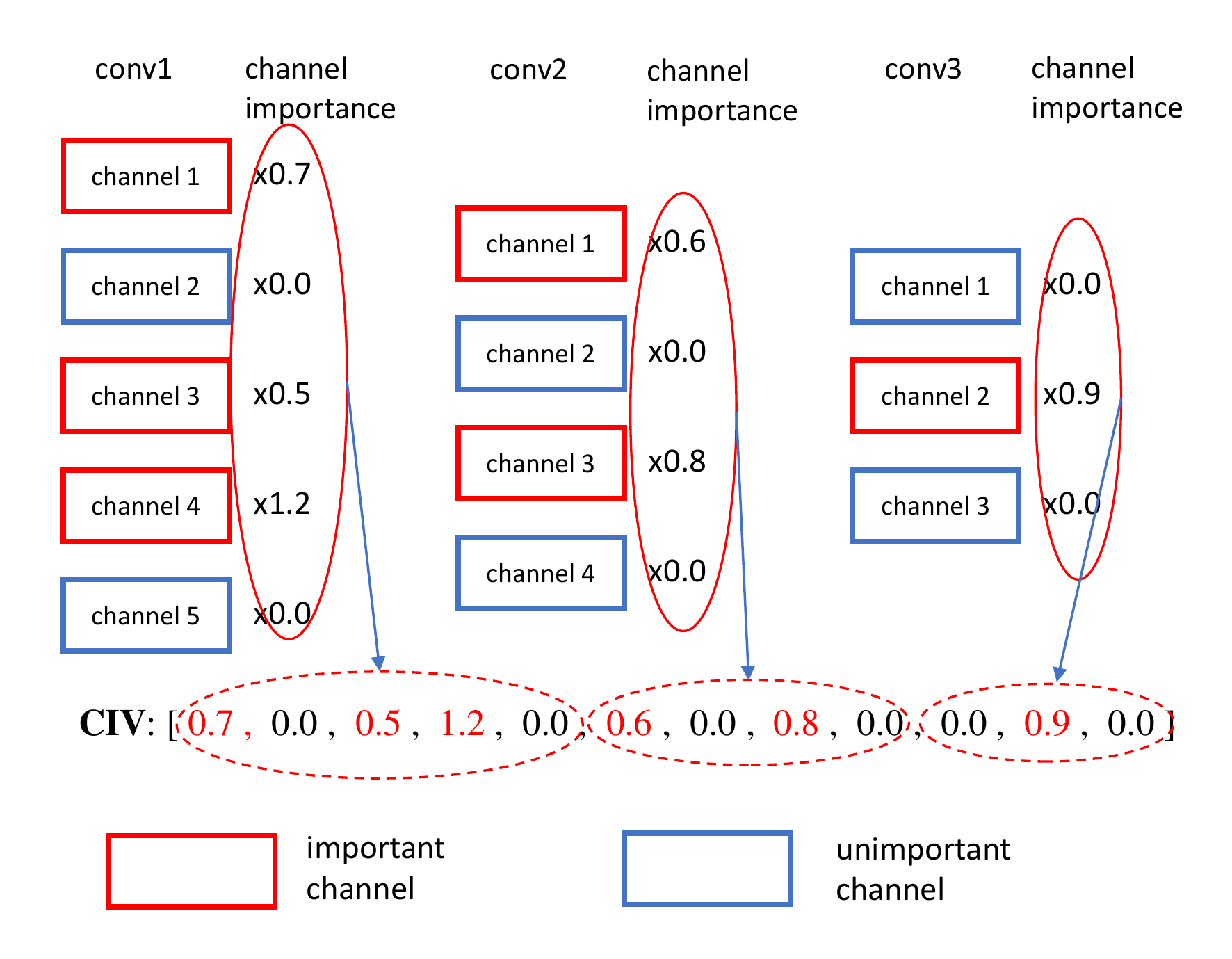}}
	\caption{Channel Importance Vector (CIV). Every CIV is a float vector. Every value in CIV indicates the importance of a channel to one certain category. The larger number, the more important.}
	\label{procstructfig}
\end{figure}

\subsection{Reconstruction Stage}
After the dissection stage, we have the CIV for each class. In order to dynamically run the network, we need one single vector to decide which channels to run. This single vector is called Combined Channel Importance Vector (CCIV), and the process which combines CIVs to get CCIV is called reconstruction.

For class set $C$, we use $\Lambda^C$ to represent its CCIV. The vector $\Lambda^C$ should be a 0-1 vector, where 1 represents we run this channel for this sub-task and vise versa.

We have tried a lot of methods to get $\Lambda^C$, such as union set, cross set and XOR set etc. We find that union set can run least channels while maintaining satisfying accuracy. In addition, XOR set can also do well in two-classes classification tasks but it is not easy to be applied on three or more classes tasks. We only introduce union set and XOR set in the following sections. They are simple but really efficient.

\subsubsection{Union Set}
\begin{figure}[t]
	\centerline{\includegraphics[width=3.0in]{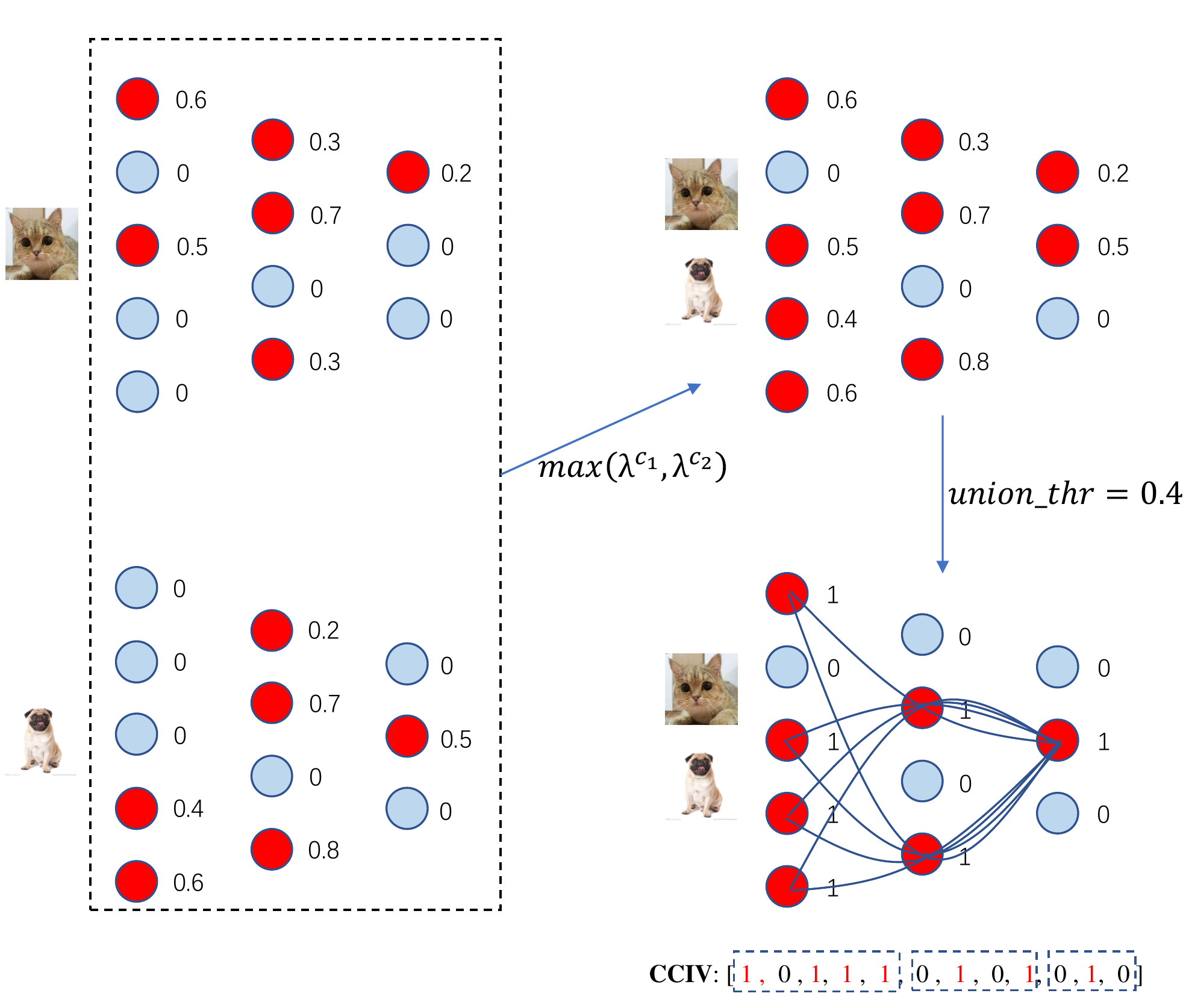}}
	\caption{Combined Channel Importance Vector (CCIV) with union set. If we need to differ cats from dogs, we run all channels that are important to them. The left two figures show the important channels for cats and dogs. For each channel, we calculate the maximum value of two CIVs, then we mask it by a threshold named $union\_thr$. In that way, we get CCIV for cat-dog sub-task.}
	\label{procstructfig}
\end{figure}

The most efficient method of getting $\Lambda^C$ is to union all $\lambda^c$s of classes in need. The motivation is easy to understand: if we want a network to classify several classes, we just need to union their important channels. Figure 3 shows the basic idea of union set.

Because $\lambda^c$s are continuous vectors, we can not use the union operation directly. To solve this problem, we propose to use the union operation of fuzzy set along with a threshold to get a 0-1 vector.

Suppose that we have a sub-task, $C$ represents the class set of it. For $\vert C \vert$ classes, we have $\vert C \vert$ $\lambda^c$s and every one of them is a continuous vector. The goal is to combine all $\lambda^c$s to generate a single vector. To get $\Lambda_j^C$ (the $j_{th}$ value of $\Lambda^C$), we first extract $\mathop{max}\limits_{c}(\lambda_j^c)$ (the maximum value among all the $j_{th}$ values of $\lambda^c$s. Then we clip the value of $\Lambda_j^C$ to 0 or 1 by a threshold named $union\_thr$. The following Formula 3 describes how we generate the $j_{th}$ value of $\Lambda^C$:

\begin{equation}
\Lambda_j^C =
\begin{cases}
1, &if \quad \mathop{max}\limits_{c}(\lambda_j^c) \ge union\_thr \\
0, &otherwise
\end{cases}
\end{equation}

Using this method, we can get a 0-1 vector $\Lambda^C$. For the dynamic neural network, $union\_thr$ is a threshold that balance the trade-off between efficiency and accuracy.

After we get $\Lambda^C$, also named CCIV, we run the original ConvNet according to it. If the value of a channel in $\Lambda^C$ is 1, we run that channel, otherwise we silence that channel.

\subsubsection{XOR Set}
\begin{figure}[t]
	\centerline{\includegraphics[width=3.0in]{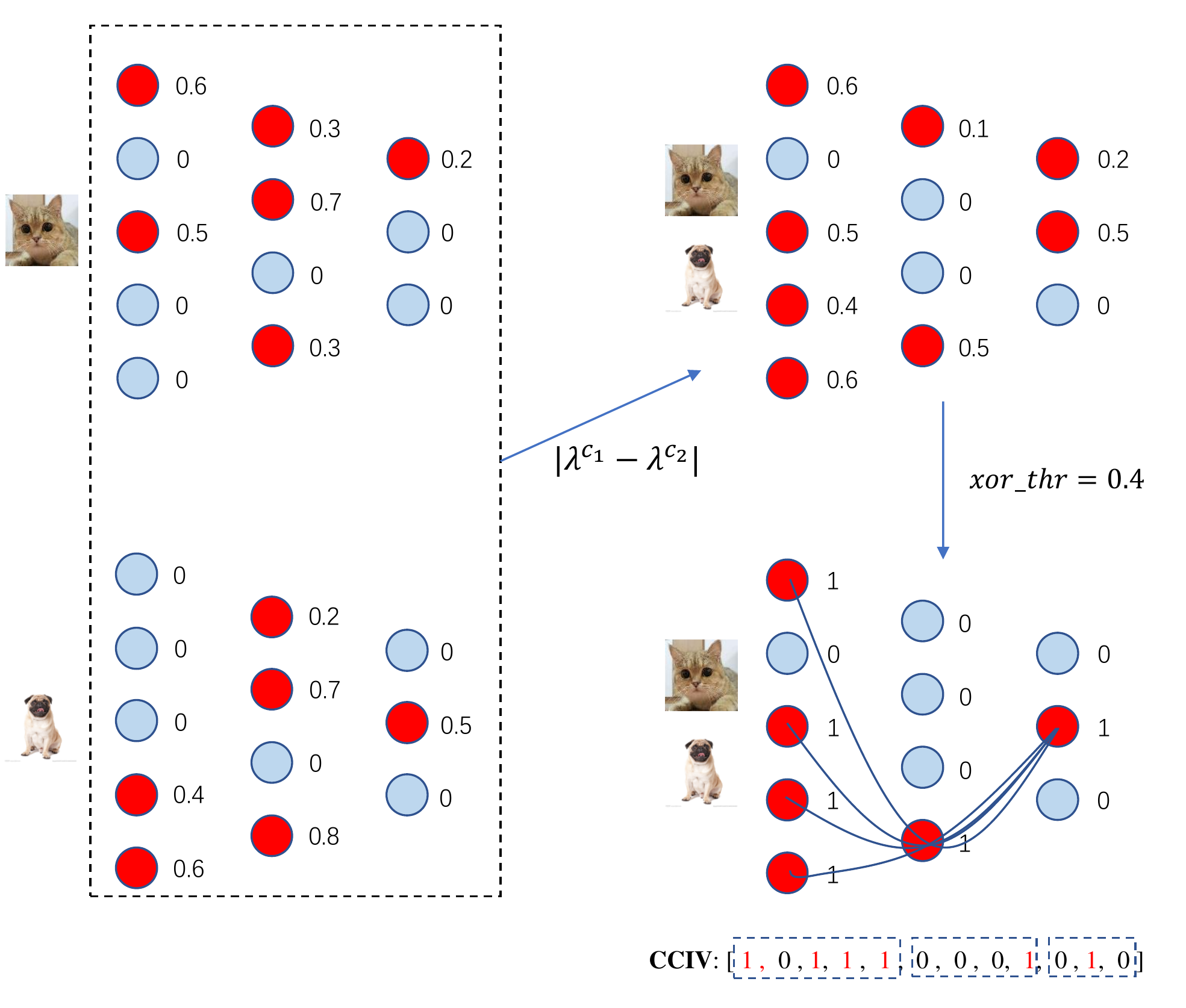}}
	\caption{Combined Channel Importance Vector (CCIV) with XOR set. We assume that common channels between cats and dogs extract similar patterns. As we aim to differ cats from dogs, we only need their different patterns. XOR set keeps the channels that have most different importance values between two classes. We first calculate the L1 distance of their CIVs, then we mask it by a threshold named $xor\_thr$ to get the final CCIV for cat-dog sub-task.}
	\label{procstructfig}
\end{figure}

The idea of XOR set comes from the semantic meanings of the $\lambda^c$s. For $i_{th}$ channel, the value $\lambda^c_i$ represents its importance for class $c$. If we only have 2 classes to classify, we only need their different features. As we know, different channels of ConvNets extract different features, so that we can use their different channels. And the different part of their $\lambda^c$s is what we call XOR set. Figure 4 shows the basic idea of XOR set.

The basic idea of XOR set is also easy to understand, but we face the same question as union set. As $\lambda^c$s are continuous vectors, we can not simply use the XOR operation to combine them. This time we define a soft-XOR operation to combine two CIVs.

Suppose that we have two CIVs $\lambda^{c_1}$ and $\lambda^{c_2}$ for class $c_1$ and class $c_2$, we need to get a 0-1 vector $\Lambda^{c_1 \oplus c_2}$. The soft-XOR operation preserves the most different channels between them. We use a threshold named $xor\_thr$ to decide which channels to keep. The $j_{th}$ value of $\Lambda^{c_1 \oplus c_2}$ can be calculated by the following formula:

\begin{equation}
\Lambda^{c_1 \oplus c_2}_j =
\begin{cases}
1, &if \quad \left\vert \lambda_j^{c_1} - \lambda_j^{c_2} \right\vert \ge xor\_thr \\
0, &otherwise
\end{cases}
,
\end{equation}

The $xor\_thr$ is also a threshold that balance the trade-off between efficiency and accuracy.

According to our experiments, the XOR set has similar performance with union set on two-classes sub-tasks. As it is not easy to apply soft-XOR operation on sub-tasks which need more than two classes, we will not show the results of XOR set in Section 4.

\section{Experiments}

\subsection{Experiment Settings}
In this paper, we use three state-of-the-art ConvNet models: VGG16, ResNet18 and MobileNetV2. VGG16 represents the classical deep ConvNets. ResNet18 is another classical deep ConvNet with skip connections. And MobileNetV2 stands for the light weight structures, which is broadly used on mobile devices. Experiments show that our method has promotions on all of them.

As for datasets, we use CIFAR-10, CIFAR-100 and ImageNet. CIFAR-10 is a light weight dataset composed of 60,000 images, 50,000 for training and 10,000 for testing. It has been divided into 10 classes and every class has 5,000 training images and 1,000 test images. Every single picture of CIFAR-10 is a 32x32 pixel RGB graph. This dataset has been broadly used for image classification and it is one of the most famous benchmark datasets in deep learning field. CIFAR-100 is very similar to CIFAR-10. The only difference is that CIFAR-100 has 100 classes. ImageNet is another famous benchmark dataset used in ILSVRC (ImageNet Large Scale Visual Recognition Challenge). It has 1.5 million RGB images with 1,000 classes. The size of each picture is not the same, but much larger than CIFAR-10. It also has a test set of 50,000 images, 50 for each class.

As for CIFAR-10 and CIFAR-100, we train VGG16, ResNet18 and MobileNetV2 with SGD optimizer. The initial learning rate, momentum and weight decay are set to 0.1, 0.9 and $1\times10^{-5}$ respectively. We train three networks for 550 epochs with batch size 256. As for data augmentation, we randomly crop the images to $32 \times 32$ after adding 4 pixels of zero padding. We also use the random horizontal flip.

As for ImageNet dataset, we use the pre-trained VGG16 (provided in torchvision) for experiment. The pre-trained model is trained on all 1,000 classes of ImageNet.

After the pre-training process, we extract the CIV for each class. For every single image, we initialize control gates vector ($\lambda$) with $1$, and iterate for $T = 30$ times. We use the SGD optimizer with learning rate $0.1$, momentum $0.9$ and no weight decay. We empirically set $\gamma=0.05$. At the end of each iteration, we clip the value of control gates to $[0, 10]$. Note that for every single image, we always maintain the top-1 prediction the same as the original network. If the top-1 prediction changes, we set all values of control gates vector to $1$, which means the network outputs exactly the same as the original network. After we get CDRPs for all images in the training set, we use $100$ CDRPs per class to generate the CIV.

We randomly chose some two-classes and three-classes sub-tasks for testing. We use union set method to combine CIVs and get CCIV for each sub-task. On CIFAR-10, we set the $union\_thr$ to $6\times10^{-3}$ for VGG16, $2\times10^{-3}$ for ResNet18 and $8\times10^{-4}$ for MobileNetV2. On CIFAR-100, we set it to $1\times10^{-2}$, $6\times10^{-3}$ and $2\times10^{-2}$ respectively. These values can be tuned to balance the trade-off between efficiency and accuracy of the dynamic networks. Then we dynamically run part of channels according to CCIV.

\subsection{Masked Softmax}
For the full-task of ConvNets, people usually use the outputs of softmax layer to represent the prediction probabilities. However, for a sub-task of ConvNet, we do not need all of these probabilities. As a result, we mask the softmax layer, w.r.t for classes not in this sub-task, the prediction probabilities for them will always be zeros. Using the masked softmax, we can exclude the effects of classes which are not considered in this sub-task.

\subsection{Results on CIFAR-10}
As the accuracy on CIFAR-10 dataset is already high enough for application, we try to run as few channels as possible to improve efficiency.

The results of VGG16 are shown in Table 1. We choose some typical classes combinations to display. The car-deer combination is the most easy pair to differentiate, and the cat-dog is easiest difficult two-class sub-task in CIFAR-10 dataset. The plane-truck has the average classification difficulty among all two-class combinations.

As for VGG16, the accuracy loss is negligible for both two-classes and three-classes sub-tasks. However, for two-classes sub-tasks, we only run 33\% channels (11.2\% parameters) on average. For three-classes sub-tasks, we only run 50\% channels (22.5\% parameters) on average. Note that there are a few cases result in great accuracy loss with the $union\_thr$ value setting described in Section 4.1, so that we need to preserve more channels through tuning the $union\_thr$ to get better performance. Cat-dog is one of these cases, and 40\% channels (14.9\% parameters) is enough for cat-dog sub-task. 

\begin{table}
	\begin{center}
		{\caption{The results of VGG16 on two and three classes sub-tasks of CIFAR-10. 10 classes stands for the full-task.}\label{table}}
		\begin{tabular}{lcccc}
			\hline
			\rule{0pt}{12pt}
			&Full-Net&Sub-Net&Running&Running\\
			sub-tasks&Acc.&Acc.&Channels&Param.
			\\
			\hline
			\\[-6pt]
			10 classes&93.98\%&93.98\%&100\%&14.73M(100\%)\\
			\\[-6pt]
			\cline{1-5}
			\\[-6pt]
			plane-truck&99.00\%&98.85\%&33\%&1.65M(11.2\%)\\
			car-deer&99.85\%&99.40\%&32\%&1.56M(10.6\%)\\
			cat-dog&92.85\%&92.20\%&40\%&2.20M(14.9\%)\\
			\\[-6pt]
			\cline{1-5}
			\\[-6pt]
			plane-bird-cat&96.70\%&95.87\%&50\%&3.31M(22.5\%)\\
			car-deer-frog&99.43\%&99.13\%&47\%&3.01M(20.4\%)\\
			dog-horse-truck&97.67\%&97.37\%&51\%&3.43M(23.3\%)\\
			\\[-6pt]
			\hline
			\\[-6pt]
		\end{tabular}
	\end{center}
\end{table}

For ResNet18, the results are shown in Table 2. We can see from the results that running channels rate of ResNet18 are a little higher than VGG16, but dynamic ResNet18 still silences $3/4$ parameters in most cases of two-classes sub-tasks. The cat-dog sub-task represents one of the very few cases that we need to preserve more parameters by tuning down $union\_thr$ to ensure the performance. As for three-classes sub-tasks, all of them need to run less than $50\%$ parameters.

\begin{table}
	\begin{center}
		{\caption{The results of ResNet18 on two and three classes sub-tasks of CIFAR-10. 10 classes stands for the full-task.}\label{table}}
		\begin{tabular}{lcccc}
			\hline
			\rule{0pt}{12pt}
			&Full-Net&Sub-Net&Running&Running\\
			sub-tasks&Acc.&Acc.&Channels&Param.
			\\
			\hline
			\\[-6pt]
			10 classes&95.90\%&95.90\%&100\%&11.18M(100\%)\\
			\\[-6pt]
			\cline{1-5}
			\\[-6pt]
			plane-truck&99.40\%&98.10\%&53\%&2.82M(25.2\%)\\
			car-deer&99.90\%&98.10\%&52\%&2.68M(24.0\%)\\
			cat-dog&93.65\%&92.25\%&73\%&5.73M(51.3\%)\\
			\\[-6pt]
			\cline{1-5}
			\\[-6pt]
			plane-bird-cat&97.53\%&96.77\%&72\%&5.34M(47.8\%)\\
			car-deer-frog&99.60\%&98.47\%&68\%&4.71M(42.1\%)\\
			dog-horse-truck&98.90\%&97.93\%&72\%&5.45M(48.7\%)\\
			\\[-6pt]
			\hline
			\\[-6pt]
		\end{tabular}
	\end{center}
\end{table}

The results of MobileNetV2 are shown in Table 3. In this table, we only display the running channels for last two convolutional layers, because the most computational-dense part of MobileNetV2 is the tail part\cite{mobilenetv3}.

We can tell from the results that the accuracy losses are still negligible. We only run 12\% channels of the 320 convolutional layers on average both in two and three classes sub-tasks. As for 1280 channels convolutional layer, the two-classes tasks only run less than 50\% channels and the three-classes tasks only run a little more than 50\% channels.

\begin{table}
	\begin{center}
		{\caption{The results of MobileNetV2 on two and three classes sub-tasks of CIFAR-10. 320 and 1280 stand for the last two convolutional layers in MobileNetV2. 10 classes stands for the full-task.}\label{table}}
		
		\begin{tabular}{lcccc}
			\hline
			\rule{0pt}{12pt}
			&Full-Net&Sub-Net&Running&Running\\
			sub-tasks&Acc.&Acc.&Channels(320)&Channels(1280)
			\\
			\hline
			\\[-6pt]
			10 classes&93.38\%&/&100\%&100\%\\
			\\[-6pt]
			\cline{1-5}
			\\[-6pt]
			plane-truck&99.00\%&98.70\%&13\%&45\%\\
			car-deer&99.90\%&99.70\%&14\%&47\%\\
			cat-dog&92.30\%&91.85\%&9\%&35\%\\
			\\[-6pt]
			\cline{1-5}
			\\[-6pt]
			plane-cat-bird&98.23\%&97.77\%&15\%&62\%\\
			bird-cat-deer&95.53\%&93.47\%&9\%&57\%\\
			deer-dog-frog&97.50\%&96.17\%&12\%&51\%\\
			\hline
			\\[-6pt]
		\end{tabular}
	\end{center}
\end{table}

\subsection{Results on CIFAR-100}

In this section, we will show more statistical results on CIFAR-100.

As we know, 100 classes have 4,950($C_{100}^2$) kinds of two-classes combinations and  161,700($C_{100}^3$) kinds of three-classes combinations. We randomly choose 200 kinds of two-classes and 200 kinds of three-classes combinations for evaluation. We compute the average accuracy of full networks, the average accuracy of the dynamic networks, the average accuracy drop, and the average running channels of the dynamic networks. The results are shown in Table 4 and Table 5.

\newcommand{\tabincell}[2]{\begin{tabular}{@{}#1@{}}#2\end{tabular}}
\begin{table}
	\begin{center}
		{\caption{The average results on 200 kinds of two-classes sub-tasks of CIFAR-100 dataset. 320 and 1280 stand for the last two convolutional layers in MobileNetV2.}\label{table}}
		\begin{tabular}{lccccc}
			\hline
			\rule{0pt}{12pt}
			&Full-Net&Sub-Net&Avg Acc.&Avg Running\\
			ConvNets&Avg Acc.&Avg Acc.&Drop&Channels
			\\
			\hline
			\\[-6pt]
			VGG16&98.15\%&97.12\%&1.04\%&48.6\%\\
			ResNet18&98.93\%&97.49\%&1.44\%&63.1\%\\
			MobileNetV2&98.84\%&98.31\%&0.54\%&\tabincell{c}{36.0\%(320)\\12.7\%(1280)}\\
			\hline
			\\[-6pt]
		\end{tabular}
	\end{center}
\end{table}

\begin{table}
	\begin{center}
		{\caption{The average results on 200 kinds of three-classes sub-tasks of CIFAR-100 dataset. 320 and 1280 stand for the last two convolutional layers in MobileNetV2.}\label{table}}
		\begin{tabular}{lccccc}
			\hline
			\rule{0pt}{12pt}
			&Full-Net&Sub-Net&Avg Acc.&Avg Running\\
			ConvNets&Avg Acc.&Avg Acc.&Drop&Channels
			\\
			\hline
			\\[-6pt]
			VGG16&96.31\%&94.82\%&1.49\%&54.92\%\\
			ResNet18&97.70\%&96.17\%&1.53\%&74.5\%\\
			MobileNetV2&98.19\%&97.70\%&0.50\%&\tabincell{c}{49.4\%(320)\\18.4\%(1280)}\\
			\hline
			\\[-6pt]
		\end{tabular}
	\end{center}
\end{table}

In our experiments, the original accuracies of VGG16, ResNet18 and MobileNetV2 on CIFAR-100 dataset are 73.8\%, 76.4\% and 75.41\% respectively. Table 4 displays the results of two-classes sub-tasks. The average accuracy of full networks is nearly 99\%. The dynamic VGG16 has 1.04\% accuracy loss on average and the dynamic ResNet18 gets 1.44\% average accuracy loss. However, the average running channels of the ResNet18 on two-classes sub-tasks is greater than VGG16. That means the ResNet18 on this dataset is more difficult to be dissected. As for the MobileNetV2, it only has 0.54\% accuracy loss on average with running channels of 36\% and 12.7\% on the last two convolution layers.

Table 5 shows the results of three-classes sub-tasks. The average accuracy loss is similar to the two-classes sub-tasks, but the running channels rate of all three ConvNets are higher than the two-classes sub-tasks. These two experiments show that our methods still have effect on CIFAR-100, and can efficiently run different part of ConvNets dynamically.

\subsection{Results on ImageNet}
For two-classes and three-classes sub-tasks on ImageNet, VGG16 has only 60\% to 80\% accuracies, which is far from enough for real-time application. To tackle this problem, we try to optimize the accuracy instead.

We randomly choose 10 classes from ImageNet to generate their CIVs. So that there are 45($C_{10}^2$) kinds of two-classes sub-tasks and 120($C_{10}^3$) kinds of three-classes sub-tasks. Then we randomly choose 40 kinds of two or three sub-tasks from these 10 classes for evaluation. As we know, we can balance the trade-off between efficiency and accuracy through tuning $union\_thr$. When generating CCIVs, we tune the $union\_thr$ so that we can preserve more than 90\% channels.

\begin{figure}[t]
	\centerline{\includegraphics[width=3.0in]{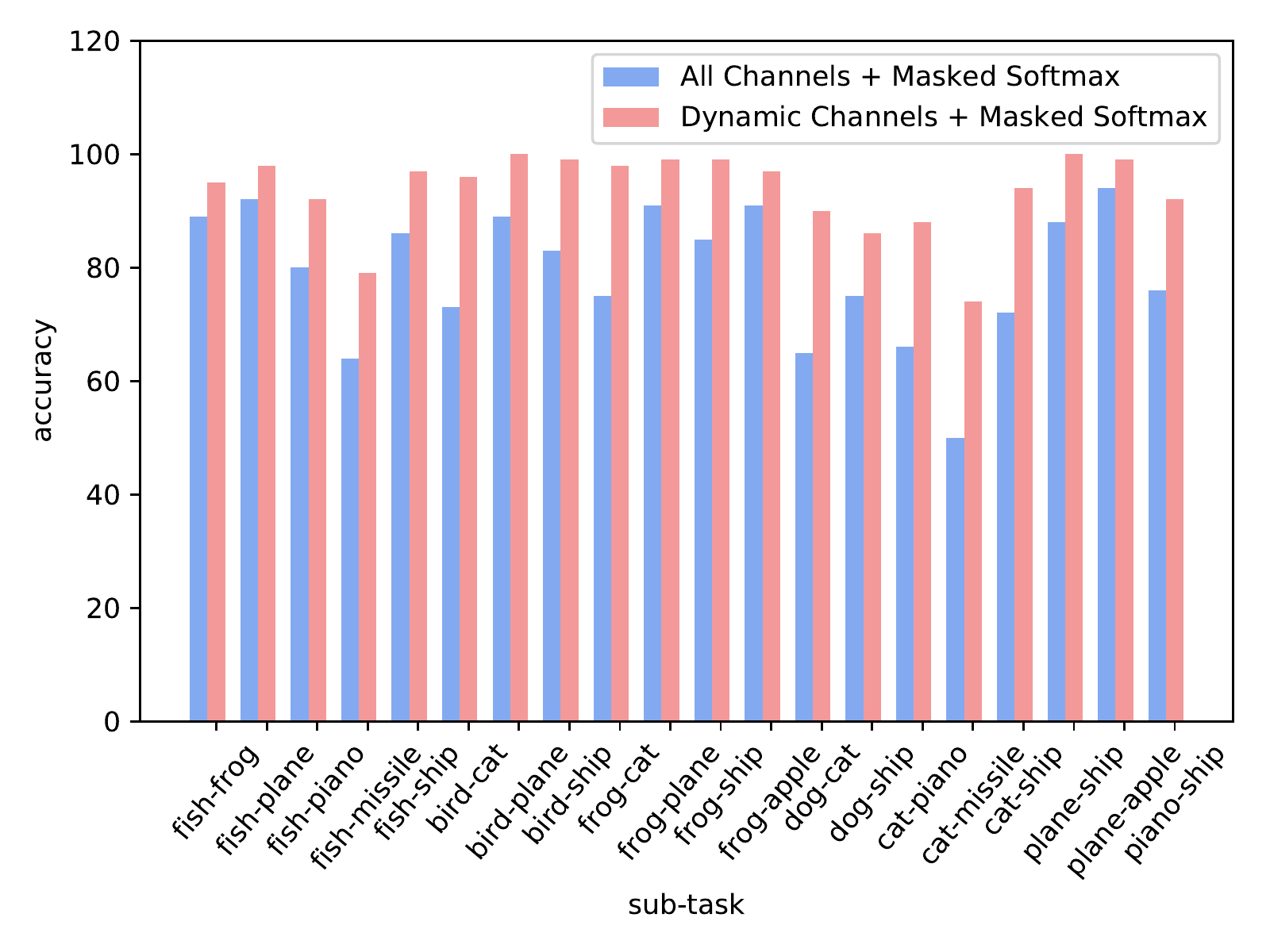}}
	\caption{Accuracy of VGG16 on two-classes sub-tasks of ImageNet.}
	\label{procstructfig}
\end{figure}

\begin{figure}[t]
	\centerline{\includegraphics[width=3.0in]{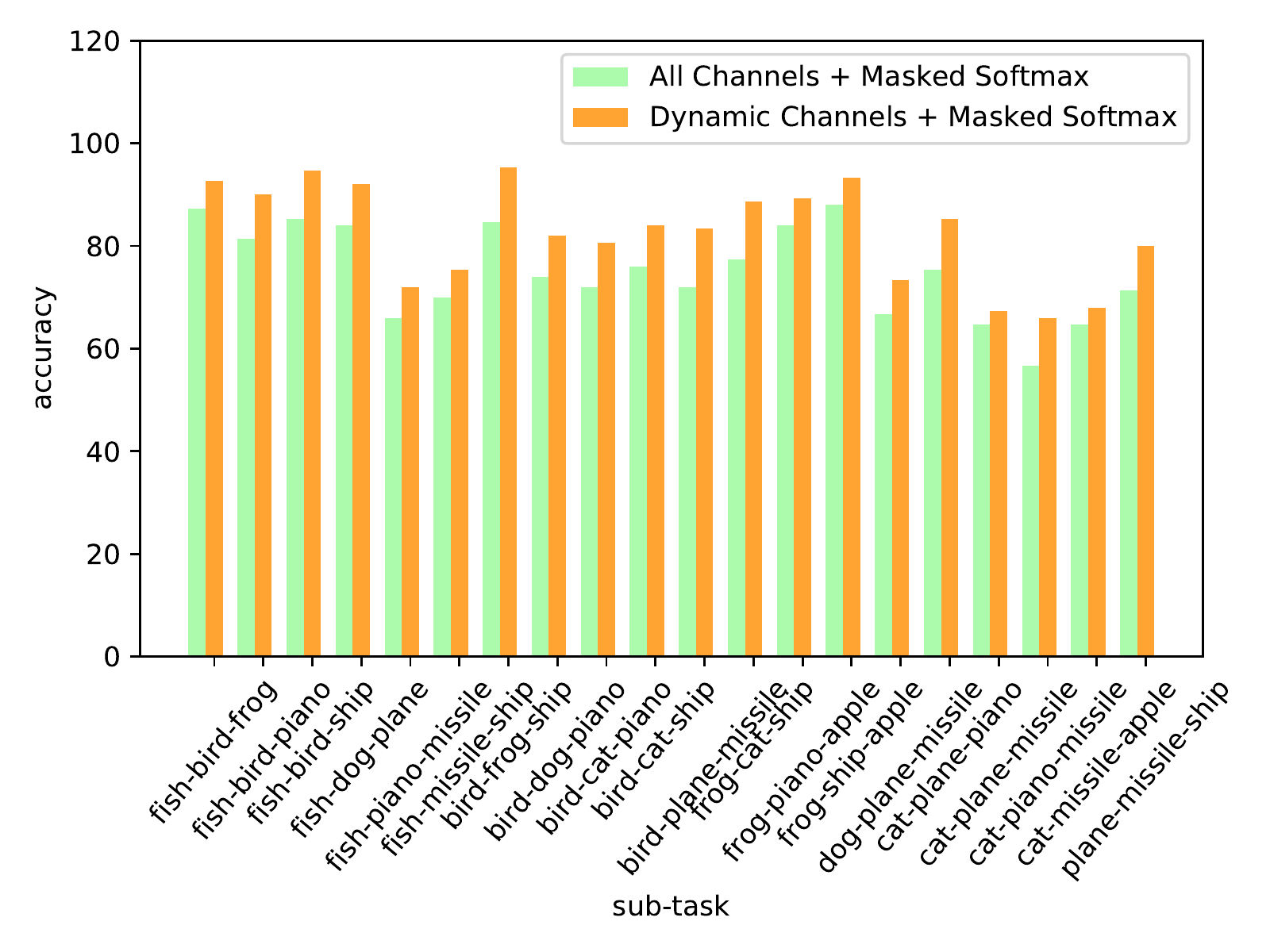}}
	\caption{Accuracy of VGG16 on three-classes sub-tasks of ImageNet.}
	\label{procstructfig}
\end{figure}

\begin{table}
	\begin{center}
		{\caption{The max and average accuracy promotions of dynamic inference VGG16 for two and three classes sub-tasks on ImageNet dataset. }\label{table}}
		\begin{tabular}{lccc}
			\hline
			\rule{0pt}{12pt}
			&Max Acc.&Avg Acc.&Min Acc.\\
			sub-tasks&Promotion&Promotion&Promotion\\
			\hline
			\\[-6pt]
			two-classes&27.0\%&14.29\%&-2.00\%\\
			three-classes&14.7\%&7.43\%&-0.67\%\\
			\hline
			\\[-6pt]
		\end{tabular}
	\end{center}
\end{table}

The results are shown in Figure 5 and Figure 6. We can see that accuracy of all sub-tasks are significantly improved by running part of channels according to their CCIVs. Table 6 shows the maximum, minimum and average accuracy promotion for two and three classes sub-tasks. The maximum and average accuracy promotions are very significant in both two and three classes sub-tasks.

From the table, we can see that the minimum accuracy promotion is negative. Note that the negative result is very few in all of our experiments. If we meet this kind of sub-tasks in real-time application, we can straightly run all channels instead.

\section{Analysis}

\subsection{Mechanism of Our Method}
In this paper, we propose CIV and CCIV, and get performance promotion by dynamically running part of channels for sub-tasks. In our point of view, the class set of a sub-task is a kind of prior knowledge. Currently, there are very few methods can apply prior knowledge on deep learning model. Our method propose a new way of applying prior knowledge on sub-tasks of ConvNets.

Why running fewer channels can promote the performance of sub-tasks? We explain this by the context information of the dataset. The context information is, for example, if the background of a image is the blue sky, it is more likely to be a plane or a bird. As we know, ImageNet has 1,000 classes, and there are many context information among all of these classes. The full network pre-trained on ImageNet contains all context information of these classes. However, if we use the full network to deal with a sub-task, the context information learned from other classes will affect the final decision.

As we have the prior knowledge, we can make use of it to get better performance. According to our CCIV, we do not run all channels and only run channels that are useful for classes of the sub-task. In that way, we use the context information more targeted. Finally, the sub-network focus on the sub-task, and it gets higher accuracy reasonably.

\subsection{Distribution of Running Channels}

\begin{figure}[t]
	\centerline{\includegraphics[width=3.0in]{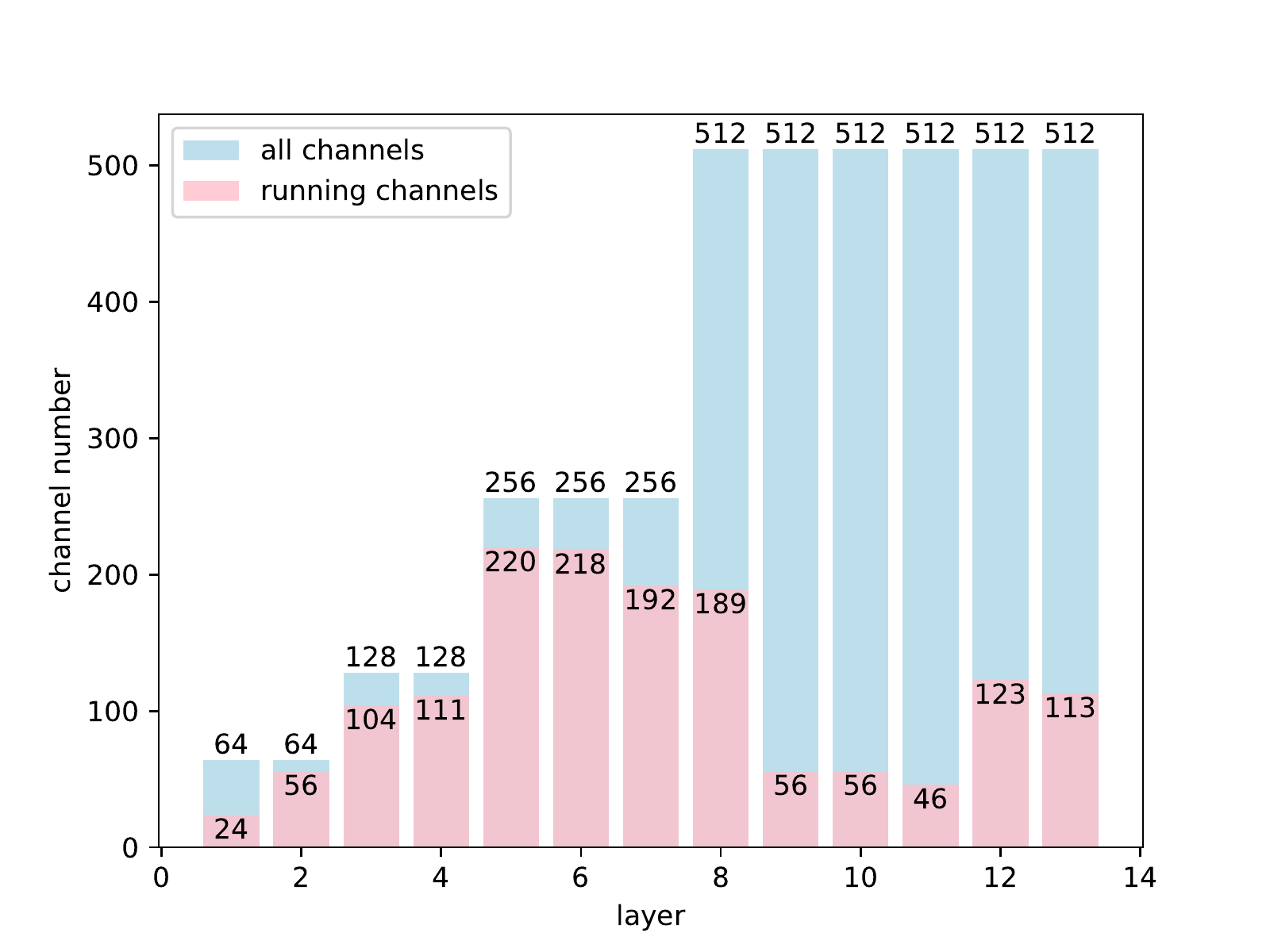}}
	\caption{The layer-wise running channels of VGG16 on plane-truck sub-task of the CIFAR-10 dataset. During the inference time, we need to run more channels in front layers, and fewer channels in high level layers. This is consistent with the fact that lower layers extract more basic features, and the higher layer extract more semantic features.} \label{procstructfig}
\end{figure}

For VGG16 on CIFAR-10 dataset, Figure 7 shows the ratio of running channels among layers. The results are from plane-truck sub-task, and we use it as an example. As for other sub-tasks, they have similar results.

As we know, the front layers of ConvNets usually extract basic features, and the higher layers extract more concrete semantic information. As a result, the channels of higher layers should be more sensitive to different categories than the front layers.

In Figure 7, the blue columns stand for the original number of channels, and the pink columns represent the running channel numbers in this sub-task. The result shows a basic trend: along with the rising of layer number, the running channel number decreases. It means that higher layers are more 'dynamic' than front layers according to our experiments. This phenomenon indicates that CIVs and CCIVs actually capture some semantic meanings of channels.

\subsection{Class-Wise Similarity Defined by CIVs}

\begin{figure}[t]
	\centerline{\includegraphics[width=3.0in]{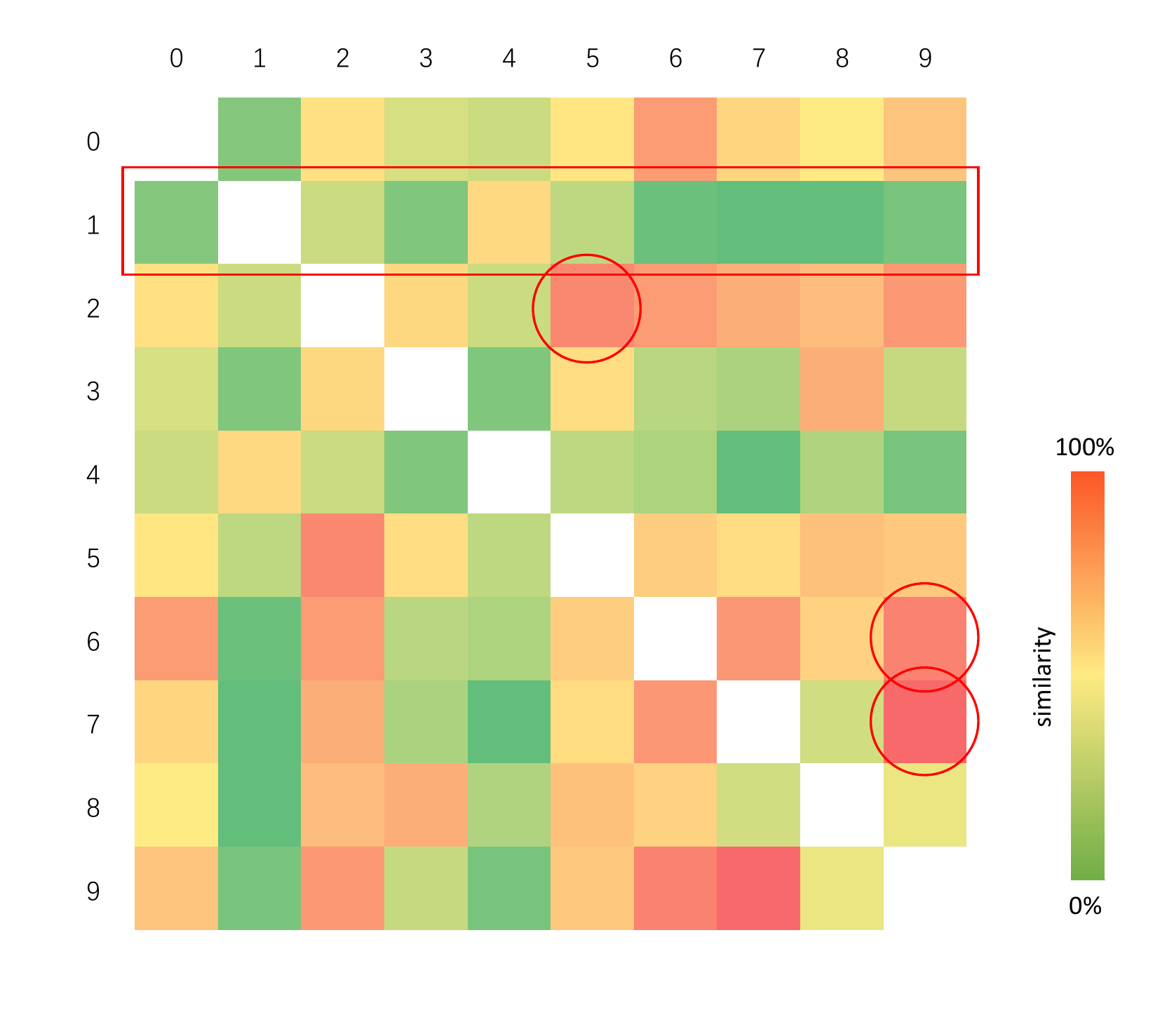}}
	\caption{The similarity heat map among categories of MNIST dataset. The similarity between class $c_1$ and class $c_2$ is defined by the IoU of their important channel sets. The red rectangle indicates that '1' is different from all other numbers. Three red circle mean that '2' is similar to '5', '6' is similar to '9', and '9' is similar to '7'. We can see that numbers with similar patterns have similar important channels. See Figure 9 for detail.}
	\label{procstructfig}
\end{figure}

As the CIVs capture some semantic meanings of channels, we speculate that similar classes will have similar important channels. As a result, we draw a heat map to show the class-wise similarities defined by CIVs. However, it is not easy to define intuitive similarity on dataset such as ImageNet, because the features of this dataset are too complex. As a result, we use a 5-layer ConvNet pre-trained on MNIST dataset to generate the CIVs. Note that we do not use any data augmentation on MNIST, so that the ConvNet extracts features directly from the original dataset.

For every class, we first define the important channel set as the set of channels that have nonzero values in CIV. Then the similarities of class $c_1$ and class $c_2$ can be defined as the $IoU$ of their important channel sets. Formula 5 shows the definition of similarity. $S_{c_i}$ represents the important channel set of class $c_i$. We suppose that similar classes will have similar CIVs.

\begin{equation}
similarity_{c_1, c_2} = \frac{\vert S_{c_1} \bigcap S_{c_2} \vert}{\vert S_{c_1} \bigcup S_{c_2} \vert}
\end{equation}

\begin{figure}[t]
	\centerline{\includegraphics[width=2.5in]{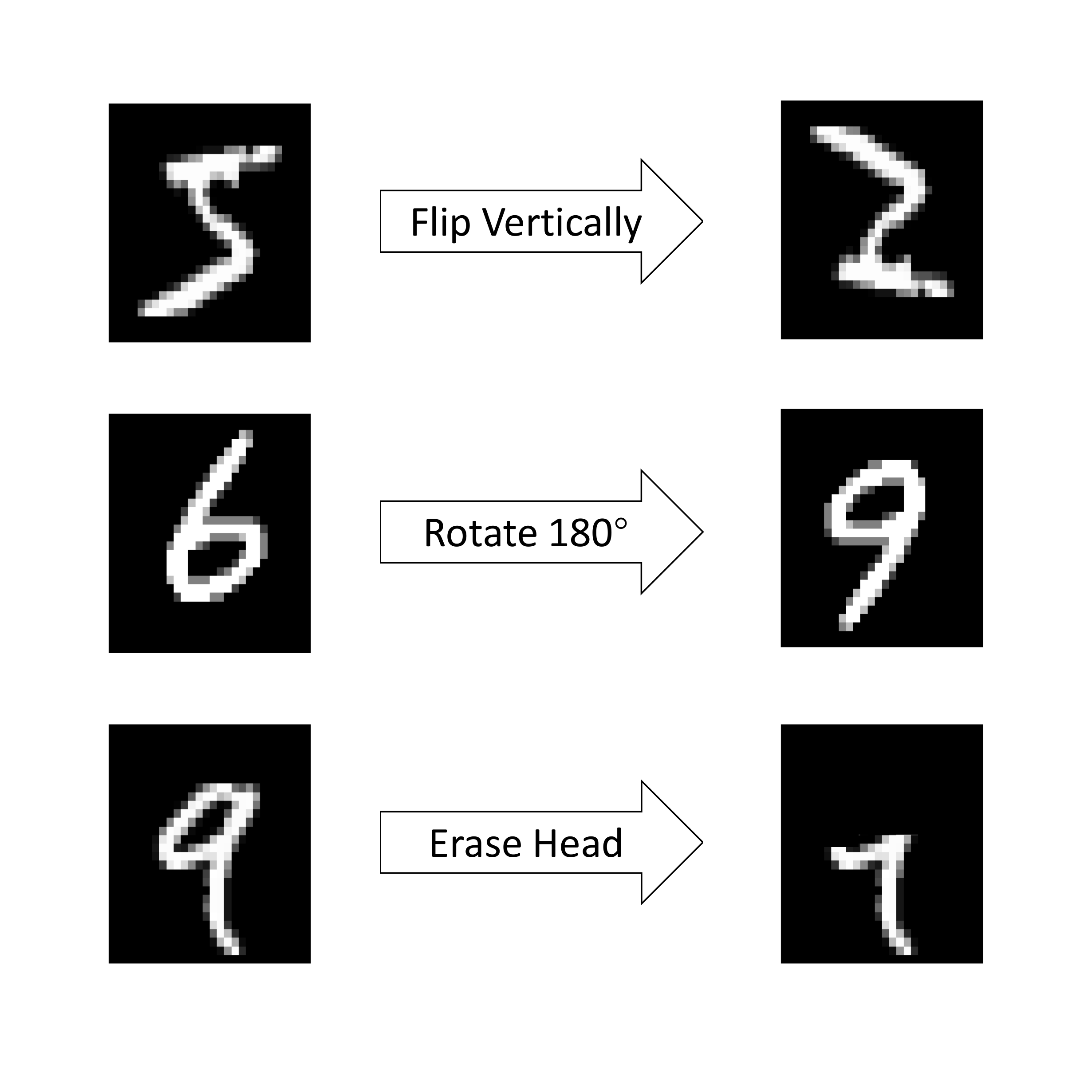}}
	\caption{Similarity of numbers. As is indicated in the heat map, '5' is similar to '2', '6' is similar to '9', '9' is similar to '7'. Numbers with high channel similarity can be easily transformed to each other. This phenomenon means that they share similar semantic patterns. And this result is another evidence to prove that our CIVs capture some semantic meanings of channels.} \label{procstructfig}
\end{figure}

Figure 8 shows the heat map. The red rectangle in the heat map shows that the handwriting '1' is not similar to any other handwriting numbers. It is in accord with our common sense. The three red circle tell us that '2' is similar to '5', '6' is similar to '9' and '9' is similar to '7'. Figure 9 explains why they are similar. If we flip a '5' vertically, we will get a '2'. If we rotate a '6' for 180 degree, we will get a '9'. If we just erase the 'head' of a '9', we will get '7'. This phenomenon is very interesting. The results indicate that if two classes share similar patterns, they will have similar CIVs, so that they have similar important channels. It is another evidence which proves that CIVs capture some semantic meanings of channels.

As the CIVs contains the semantic meanings, we use it to get CCIVs and dynamically run the network for sub-tasks. Partly running the network for sub-tasks prevents the behavior of ConvNet from being influenced by the unimportant channels. In that way, the dynamic ConvNet focuses on the sub-task, and uses the context information more targeted.

\subsection{Limitation and Future Work}

Although the results in some of our experiments are remarkable, our method still has its limitations.

Firstly, on CIFAR-10 and CIFAR-100, dynamic ResNet18 usually perform worse than dynamic VGG16. As we know, ResNet has skip connections, this indicates that our method has difficulty when handling with skip connections. We need to pay more attention to skip connections in the future work.

In addition, although our method gains remarkable accuracy promotion on ImageNet, it only reduces 10\% channels on average. For future work, we can design a better structure from scratch so that it can run less channels for any sub-task.

Moreover, although our method capture some semantic meanings of channels, it is still far from real interpretability. There are still a lot of works remain to be done. In the future, we need to shed more light on the black box of neural networks.

\section{Conclusion}

In this paper, we propose to dynamically run different channels of ConvNets for different sub-tasks. We extract the most important channels and get a Channel Importance Vector (CIV) for every single class, then we propose two efficient methods to merge these CIVs for any sub-task. The merged vector is called Combined Channel Importance Vector (CCIV). Then we dynamically run channels according to the CCIVs of sub-tasks. For VGG16 pre-trained on small dataset (CIFAR-10), we only run 11\% parameters on average for two-classes sub-tasks with negligible accuracy loss. For VGG16 trained on large-scale dataset (ImageNet), our method gains 14.29\% accuracy promotion for two-class sub-tasks on average, which is remarkable. In addition, analysis show that our method actually captures some semantic meanings of channels, and uses context information more targeted for sub-tasks of ConvNets.

\bibliography{ecai}
\end{document}